\documentclass[conference]{IEEEtran}
\IEEEoverridecommandlockouts

\usepackage{cite}
\usepackage{amsmath,amssymb,amsfonts}
\usepackage{graphicx}
\usepackage{textcomp}
\usepackage{xcolor}
\usepackage{booktabs}
\usepackage{multirow}
\usepackage[ruled,linesnumbered]{algorithm2e}
\usepackage{enumitem}

\title{Confidence-Based Adaptive Retrieval Fails Where It Matters Most: A Paradox in Biomedical Question Answering%
\thanks{This work was supported by the National Science and Technology Major Project (2024ZD0522400).}
}

\author{
\IEEEauthorblockN{Chengying She\IEEEauthorrefmark{1}, Siyu Deng\IEEEauthorrefmark{2}, Haiyan Ge\IEEEauthorrefmark{2}, Lizhuang Liu\IEEEauthorrefmark{1}}
\IEEEauthorblockA{\IEEEauthorrefmark{1}\textit{Shanghai Advanced Research Institute, Chinese Academy of Sciences}\\
\textit{University of Chinese Academy of Sciences}, Shanghai, China\\
shechengying2024@sari.ac.cn, liulz@sari.ac.cn\\
\IEEEauthorrefmark{2}\textit{Department of Respiratory and Critical Care Medicine, Huadong Hospital}\\
\textit{Fudan University}, Shanghai, China\\
dengsiyu021012@163.com, haiyan\_ge@fudan.edu.cn}
\thanks{Haiyan Ge and Lizhuang Liu are the corresponding authors.}
}

\begin{document}

\maketitle

\begin{abstract}
Large language models (LLMs) are increasingly used for biomedical question answering, but plausible incorrect answers remain a safety risk in clinical settings. Retrieval-augmented generation (RAG) can ground responses in medical literature, while confidence-based adaptive RAG retrieves only when the model appears uncertain. We evaluate this routing strategy on three biomedical QA benchmarks of increasing difficulty and compare models across a broad range of parameter scales. The results reveal a paradox: confidence is well calibrated (AUROC 0.78--0.89) on standard medical exam questions where retrieval adds little value (+2.4\% to -1.4\%), yet becomes \emph{near-random} (AUROC 0.506) on specialist-level questions where retrieval helps most (+11.2\%). Five uncertainty signals---first-token probability, margin, entropy, consistency-based sampling, and verbalized confidence---show the same failure pattern. Five post-hoc calibration correction strategies, including density-ratio reweighting and rank-preserving transformation, do not recover a useful routing signal on hard questions (best AUROC: 0.507). Cross-model comparison shows that larger parameter scale and higher base accuracy do not guarantee reliable confidence-based routing under specialist-domain difficulty shift. These findings challenge the premise of confidence-based adaptive RAG: in high-stakes biomedical settings, defaulting to always-retrieve is safer and more accurate than confidence-based routing when calibration on the target domain is unknown.
\end{abstract}

\begin{IEEEkeywords}
biomedical question answering, retrieval-augmented generation, adaptive retrieval, confidence calibration, clinical AI safety
\end{IEEEkeywords}

\section{Introduction}

Clinical decision support systems (CDSS) have long relied on knowledge-based retrieval to assist physicians in diagnosis and treatment planning \cite{topol2019high, greenes2017clinical}. Large language models (LLMs) have transformed biomedical question answering and now achieve expert-level performance on medical licensing exams \cite{singhal2023large}. However, they still generate plausible but factually incorrect responses, a critical liability when clinical errors can affect patient safety \cite{topol2019high}.

\begin{figure}[t]
  \centering
  \includegraphics[width=\columnwidth]{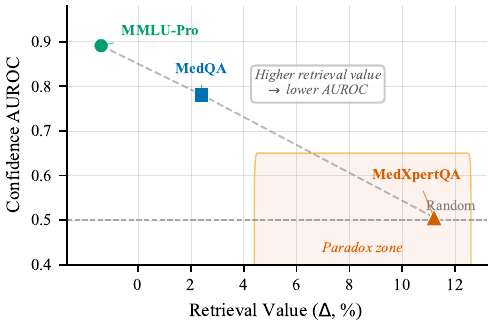}
  \caption{The confidence--retrieval paradox across three biomedical QA benchmarks. Retrieval value ($\Delta$, x-axis) and confidence AUROC (y-axis) are inversely correlated. On MedXpertQA (red), retrieval provides the largest gain yet confidence is near-random.}
  \label{fig:paradox}
\end{figure}

Retrieval-augmented generation (RAG) addresses this by grounding model outputs in authoritative biomedical literature \cite{lewis2020retrieval, gao2024retrieval}, effectively combining the reasoning capabilities of LLMs with up-to-date medical knowledge. Yet retrieval is not universally beneficial: it can introduce noise, increase latency, and even hurt performance through knowledge conflicts when retrieved content contradicts the model's parametric knowledge \cite{choi2025conflict}. This has motivated \emph{adaptive RAG}---selectively retrieving only when the model needs help, using its own confidence as the routing signal \cite{chen2025first, yao2025seakr, jiang2023flare, jeong2024adaptive, asai2024selfrag}.

The core assumption behind adaptive RAG is that low-confidence questions are precisely those where retrieval helps most. Prior work suggests LLMs ``mostly know what they know'' \cite{kadavath2022language}. Yet this assumption remains largely untested under distribution shift---the scenario most relevant to clinical practice, where physicians encounter questions spanning from routine to specialist difficulty. In biomedical settings, the cost of failing to retrieve when needed (missed diagnostic information) can far exceed the cost of unnecessary retrieval (added latency), making the reliability of confidence-based routing a patient safety concern.

We evaluate confidence-based adaptive RAG across three biomedical QA benchmarks spanning standard medical exams to expert-level specialist questions. Our central finding is a paradox: \emph{confidence-based routing works where retrieval is least needed and fails where it matters most}. On standard questions (72--81\% no-retrieval accuracy), confidence is well calibrated but retrieval provides negligible gains. On specialist questions (15\% no-retrieval accuracy), retrieval offers a $+11.2\%$ improvement, yet confidence is near-random (AUROC 0.506), causing adaptive routing to capture almost none of the available gain. Five post-hoc correction strategies---including methods from domain-shift-aware conformal prediction \cite{lin2025dscp} and rank-preserving calibration \cite{hu2026rapcal}---do not resolve this failure (best AUROC: 0.507), indicating signal absence rather than simple miscalibration. We further compare open-source models across different parameter scales, finding that the paradox persists across model families and sizes. Fig.~\ref{fig:paradox} illustrates this inverse relationship between retrieval value and confidence calibration.

\section{Related Work}

\subsection{Biomedical Question Answering}
Biomedical QA has evolved from rule-based systems to neural approaches trained on curated medical corpora. BioASQ \cite{tsatsaronis2015bioasq} established a benchmark for biomedical semantic indexing and QA, while recent LLMs have achieved expert-level performance on medical licensing exams \cite{singhal2023large}. Reliability on specialist-level questions remains limited \cite{zuo2025medxpertqa}; Jin et al.~\cite{jin2022biomedical} highlight the persistent gap between standard benchmarks and real-world clinical complexity. Retrieval effectiveness also varies across medical datasets \cite{xiong2024benchmarking}, and retrieved evidence can conflict with parametric knowledge \cite{choi2025conflict}. Our contribution is distinct: we show that adaptive systems fail to retrieve \emph{when it would help most}, not merely that retrieval can hurt.

\subsection{Adaptive Retrieval}
Prior adaptive RAG methods include FLARE \cite{jiang2023flare}, which retrieves when generated tokens have low probability, Self-RAG \cite{asai2024selfrag}, which learns retrieval and critique tokens, and Adaptive-RAG \cite{jeong2024adaptive}, which routes queries to different retrieval strategies based on question complexity. Chen et al.~\cite{chen2025first} and Yao et al.~\cite{yao2025seakr} use first-token probabilities or hidden states for retrieval decisions. Most recently, Moskvoretskii et al.~\cite{moskvoretskii2025adaptive} evaluate 35 uncertainty estimation methods across 6 QA datasets, finding that simple methods often outperform complex adaptive pipelines. Our work identifies a fundamental limitation shared by all confidence-based approaches: they fail precisely where retrieval is most valuable.

\subsection{Confidence and Calibration}
Kadavath et al.~\cite{kadavath2022language} show LLMs can assess their own knowledge, while Guo et al.~\cite{guo2017pmlr} established temperature scaling as a standard post-hoc calibration technique for neural networks. Recent work reveals that alignment (RLHF) degrades calibration \cite{xiao2025restoring}, and that calibration quality varies substantially across domains and models. These findings suggest that confidence-based routing---which assumes well-calibrated uncertainty---may be unreliable under domain shift, yet this connection has not been systematically evaluated in the context of adaptive retrieval for biomedical applications.

\subsection{RAG in Biomedical Applications}
RAG is increasingly used in biomedical and clinical NLP. Prior benchmarks show that retrieval quality and corpus fit strongly affect outcomes \cite{xiong2024benchmarking}, while knowledge conflicts between retrieved passages and parametric knowledge can mislead generation \cite{choi2025conflict}. These findings underscore that biomedical retrieval must be both relevant and reliable. Unlike work focused on \emph{when retrieval hurts}, we study \emph{when the decision to retrieve goes wrong}, a routing failure that is especially consequential in clinical settings.

\subsection{Domain Shift and Calibration Correction}
Recent work addresses calibration under distribution shift. Lin et al.~\cite{lin2025dscp} propose DS-CP, which reweights calibration samples using density ratio estimation in embedding space to maintain valid conformal prediction sets across domains. Hu et al.~\cite{hu2026rapcal} introduce RAPCal, observing that while absolute confidence fails under shift, relative rankings remain more stable; they leverage rank-preserving pseudo-label construction for cross-domain calibration. Both methods assume at least some discriminative signal exists in the confidence scores. We show that under extreme difficulty shift in biomedical QA, even the ranking becomes near-random (AUROC $\approx$ 0.5), making these correction approaches fundamentally insufficient.

\section{Method}

\subsection{Confidence Signals}

Following \cite{chen2025first}, we extract confidence from the first-token logits of the LLM's answer. For a question with $K$ choices, the model assigns probability $p_i$ to each option. We evaluate three first-token signals: \textbf{Max Probability} ($\max_i p_i$), \textbf{Margin} ($p_1 - p_2$), and \textbf{Entropy} ($-\sum_{i} p_i \log p_i$). Additionally, we evaluate two generation-based signals: \textbf{Consistency} samples $N{=}5$ answers at $T{=}0.7$ and measures the agreement rate (fraction of samples matching the greedy answer); \textbf{Verbalized Confidence} prompts the LLM to rate its certainty on a 1--5 scale, which is then mapped to $[0, 1]$. These five signals span the major categories of LLM uncertainty estimation used in prior adaptive RAG work.

\subsection{Adaptive Retrieval Algorithm}

Algorithm~\ref{alg:adaptive} summarizes the adaptive retrieval procedure. The LLM first answers without retrieval and yields a first-token margin. If the margin exceeds threshold $\tau$, the answer is returned directly; otherwise, the top-5 retrieved documents are prepended to the prompt and the LLM answers again. Lower $\tau$ values retrieve less often but risk missing helpful evidence; higher values retrieve more often but may add noise and latency.

\begin{algorithm}[t]
\SetAlgoLined
\DontPrintSemicolon
\KwIn{Question $\mathbf{q}$, options $\{o_1, ..., o_K\}$, threshold $\tau$}
\KwOut{Predicted answer $\hat{a}$}
$\hat{a}, \{p_1, ..., p_K\} \leftarrow \text{LLM}(\mathbf{q}, \{o_i\})$ \tcp*{greedy decode}
$m \leftarrow p_{(1)} - p_{(2)}$ \tcp*{margin = top-1 $-$ top-2}
\uIf{$m \geq \tau$}{
    \Return $\hat{a}$ \tcp*{high confidence, skip retrieval}
}
\Else{
    $\mathcal{D} \leftarrow \text{Retrieve}(\mathbf{q}, \text{top-}5)$\;
    $\hat{a}' \leftarrow \text{LLM}(\mathbf{q}, \{o_i\}, \mathcal{D})$\;
    \Return $\hat{a}'$\;
}
\caption{Confidence-Driven Adaptive Retrieval}
\label{alg:adaptive}
\end{algorithm}

\subsection{Calibration Correction}

The anti-predictive signal on specialist-level questions raises a natural question: can post-hoc correction methods, designed for calibration under domain shift, restore the routing signal? Drawing on recent methods from domain-shift-aware conformal prediction \cite{lin2025dscp} and rank-preserving calibration \cite{hu2026rapcal}, we propose \emph{Difficulty-Aware Calibration Correction} (DACC), which combines two complementary strategies.

\paragraph{Density Ratio Estimation}
Following \cite{lin2025dscp}, we embed all questions using Qwen3-Embedding-4B and train a logistic regression domain classifier to predict whether a question comes from the source domain (MedQA) or target domain. For each target-domain question $j$, we compute the density ratio weight:
\begin{equation}
w_j = \hat{P}(W{=}1 \mid Z{=}z_j) \, / \, (1 - \hat{P}(W{=}1 \mid Z{=}z_j))
\end{equation}
Questions that are source-like (high $w_j$) should have more reliable confidence scores.

\paragraph{Rank-Based Signal Correction}
Inspired by \cite{hu2026rapcal}, we leverage the observation that relative rankings may remain stable even when absolute values fail. We compute empirical ranks $r_i = (\text{rank}(m_i) + 0.5) / N$ and apply U-shaped stretching with optional inversion for anti-predictive signals:
\begin{equation}
p_{\text{shaped},i} = \begin{cases} 0.5(2r_i')^\alpha & \text{if } r_i' < 0.5 \\ 1 - 0.5(2(1-r_i'))^\alpha & \text{if } r_i' \geq 0.5 \end{cases}
\end{equation}
where $r_i' = 1 - r_i$ inverts the ranking when AUROC $<$ 0.5, and $\alpha$ controls stretching strength.

\paragraph{DACC Combination}
The corrected signal combines both strategies with a learnable interpolation weight $\lambda$:
$m_{\text{corrected},j} = \lambda \cdot w_j \cdot (1 - m_j) + (1 - \lambda \cdot w_j) \cdot p_{\text{shaped},j}$
where high density-ratio questions ($w_j \gg 1$) rely more on the inverted margin $(1 - m_j)$, while low-ratio questions rely more on rank correction $p_{\text{shaped},j}$. The interpolation weight $\lambda$ controls the overall influence of the density-ratio signal. Hyperparameters $\alpha \in \{1.5, 2.0, 3.0\}$ (stretching strength) and $\lambda \in \{0.3, 0.5, 0.7\}$ (interpolation weight) are swept on a held-out validation set.

\subsection{Experimental Setup}

\textbf{Model.} Qwen2.5-32B-Instruct \cite{qwen2024qwen25} for all experiments, with Qwen3-Embedding-4B for dense retrieval (top-5, cosine similarity). All inference runs use greedy decoding for confidence extraction and $T{=}0.7$ for consistency sampling.

\noindent\textbf{Datasets.} Three biomedical QA benchmarks of increasing difficulty:
\begin{itemize}[nosep]
  \item \textbf{MedQA-USMLE} \cite{jin2021medqa}: 4-choice US licensing exam questions, reflecting standard medical knowledge (500 test).
  \item \textbf{MMLU-Pro Medical} \cite{wang2024mmlupro}: 3--10 choice questions from the Health \& Biology categories, covering broader biomedical knowledge (500).
  \item \textbf{MedXpertQA-Text} \cite{zuo2025medxpertqa}: 10-choice specialist-level questions requiring expert reasoning, representing the hardest evaluation (500).
\end{itemize}

\noindent\textbf{Corpus.} 10,000 documents from MedQA training set (matched for MedQA, mismatched for others). This setup intentionally tests cross-domain retrieval to simulate the realistic clinical scenario where the retrieval corpus may not perfectly match the question distribution.

\noindent\textbf{Metrics.} We report: (1) \emph{No-RAG accuracy}---answering without retrieval; (2) \emph{Always-retrieve accuracy}; (3) \emph{AUROC} of the confidence signal for predicting answer correctness; (4) \emph{Help/Hurt} counts; and (5) \emph{Adaptive accuracy} at the best threshold $\tau$.

\noindent\textbf{Evaluation Protocol.} For each question, the LLM first answers without retrieval. We extract the first-token probability distribution over answer choices and compute confidence signals. For adaptive retrieval, we sweep the confidence threshold $\tau$ from 0 to 1 in 0.05 increments and report the best achievable adaptive accuracy. This oracle-threshold protocol upper-bounds adaptive performance; any real deployment (which must fix $\tau$ in advance) would achieve equal or lower accuracy. We use McNemar's test to assess statistical significance of Help vs.\ Hurt counts.

\section{Experiments}

\subsection{Main Results}

Table~\ref{tab:main} presents the cross-dataset comparison. Retrieval value ($\Delta$) and confidence AUROC are inversely related (Fig.~\ref{fig:paradox}).

\begin{table}[t]
  \centering
  \caption{Cross-dataset comparison of retrieval effectiveness and confidence calibration. $\Delta$ = Always $-$ No-RAG (retrieval value). AUROC = margin signal predicting answer correctness. H/H = retrieval Help/Hurt counts. The inverse relationship between $\Delta$ and AUROC reveals the paradox.}
  \label{tab:main}
  \resizebox{\columnwidth}{!}{
  \begin{tabular}{lccccc}
    \toprule
    \textbf{Dataset} & \textbf{No-RAG} & \textbf{Always} & $\boldsymbol{\Delta}$ & \textbf{AUROC} & \textbf{H/H} \\
    \midrule
    MedQA       & 72.2 & 74.6 & $+2.4$ & 0.781 & 23/17 \\
    MMLU-Pro    & 80.8 & 79.4 & $-1.4$ & 0.891 & 11/13 \\
    MedXpertQA  & 14.8 & 26.0 & $\mathbf{+11.2}$ & \textbf{0.506} & 66/13 \\
    \bottomrule
  \end{tabular}
  }
\end{table}

\paragraph{MedQA-USMLE} The LLM achieves 72.2\% without retrieval, reflecting its strong general medical knowledge. Retrieval adds a modest $+2.4\%$, with 23 questions helped and 17 hurt. Confidence is well calibrated (AUROC 0.781), meaning the model can reliably identify which questions it is likely to answer incorrectly. However, since retrieval provides only marginal gains, this well-calibrated confidence is of limited practical value for routing.

\paragraph{MMLU-Pro Medical} The LLM achieves 80.8\% without retrieval---higher than MedQA, reflecting the broader (but less specialized) nature of the questions. Retrieval \emph{hurts} overall ($-1.4\%$), with Help/Hurt (11/13) confirming that retrieval harms more than it helps. Confidence is best calibrated here (AUROC 0.891), yet this excellent calibration paradoxically means the model correctly identifies questions where it is already correct, while retrieval actively degrades the well-calibrated model's performance.

\paragraph{MedXpertQA} The paradox is starkest on this specialist-level benchmark. The model answers only 14.8\% correctly---barely above the 10\% random baseline for 10-choice questions. Retrieval delivers a substantial $+11.2\%$ improvement ($p < 0.0001$), with 66 questions helped and only 13 hurt. This is precisely the scenario where adaptive retrieval should shine. Yet confidence AUROC is 0.506: the model's confidence provides essentially no routing information. The adaptive strategy cannot distinguish between questions where retrieval would help and those where it would not, capturing almost none of the available $+11.2\%$ gain.

\subsection{Multi-Signal Ablation}

Table~\ref{tab:multisignal} compares five uncertainty signals on 200 MedXpertQA samples. Consistency uses $N{=}5$ sampled generations ($T{=}0.7$); verbalized confidence asks the LLM to rate certainty on 1--5.

\begin{table}[t]
  \centering
  \caption{Multi-signal ablation on MedXpertQA (200 samples). All signals are at or below random (AUROC 0.5). Random retrieval (19.4\%) outperforms all adaptive strategies; always-retrieve (27.0\%) provides a substantial margin.}
  \label{tab:multisignal}
  \small
  \begin{tabular}{lcc}
    \toprule
    \textbf{Signal} & \textbf{AUROC} & \textbf{Adapt. ACC} \\
    \midrule
    Max Probability     & 0.394 & 17.5\% \\
    Margin              & 0.391 & 17.5\% \\
    Entropy             & 0.390 & 17.5\% \\
    Consistency ($N{=}5$) & 0.522 & 17.5\% \\
    Verbalized (1--5)   & 0.391 & 19.0\% \\
    \midrule
    Random baseline     & --- & 19.4\% \\
    Always Retrieve     & --- & \textbf{27.0\%} \\
    \bottomrule
  \end{tabular}
\end{table}

All signals fail. First-token signals are anti-predictive (AUROC $<$ 0.5), meaning higher confidence actually correlates with \emph{incorrect} answers. Consistency ($N{=}5$) is near-random---the model is \emph{consistently wrong in the same way}, generating the same incorrect answer across all samples. Verbalized confidence is similarly anti-predictive. Notably, random retrieval (19.4\%) outperforms all adaptive strategies, and always-retrieve (27.0\%) provides a substantial 9.6\% gain over the best adaptive approach.

\subsection{Confidence Signal Comparison}

Table~\ref{tab:signals} shows AUROC for all three first-token signals across datasets. The inverse pattern is consistent across all signals: all three are well-calibrated on MedQA and MMLU-Pro (AUROC 0.77--0.89) but collapse to near-random or anti-predictive on MedXpertQA (AUROC 0.45--0.47). This uniformity suggests the failure is not specific to any single confidence extraction method but reflects a deeper limitation of LLM self-assessment under difficulty shift.

\begin{table}[t]
  \centering
  \caption{AUROC of three confidence signals. Calibration degrades as retrieval value increases.}
  \label{tab:signals}
  \small
  \begin{tabular}{lccc}
    \toprule
    \textbf{Signal} & \textbf{MedQA} & \textbf{MMLU-Pro} & \textbf{MedXpert} \\
    \midrule
    Max Prob.  & 0.78 & 0.82 & 0.46 \\
    Margin     & 0.79 & 0.83 & 0.45 \\
    Entropy    & 0.77 & 0.81 & 0.47 \\
    \bottomrule
  \end{tabular}
\end{table}

\begin{figure*}[t]
  \centering
  \includegraphics[width=0.95\textwidth]{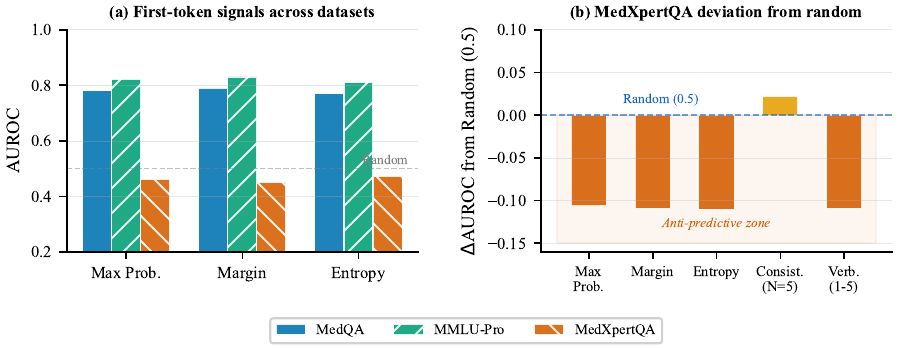}
  \caption{Visual overview of confidence signal behavior. (a) First-token signals show a consistent degradation pattern from easy to hard datasets. (b) On MedXpertQA, all signals deviate minimally from random (0.5); first-token signals are anti-predictive while consistency barely exceeds chance. Exact values in Table~\ref{tab:signals} and Table~\ref{tab:multisignal}.}
  \label{fig:signals}
\end{figure*}

\subsection{Can Calibration Correction Resolve the Paradox?}

A natural question is whether the near-random confidence signal on MedXpertQA can be corrected using methods from the calibration literature. We evaluate five correction strategies detailed above across all three datasets.

\begin{table}[t]
  \centering
  \caption{AUROC of corrected confidence signals on three datasets. No method meaningfully improves routing on MedXpertQA (best: 0.507 vs.\ 0.506 raw). On MedQA and MMLU-Pro, methods either preserve or destroy the existing signal.}
  \label{tab:correction}
  \small
  \begin{tabular}{lccc}
    \toprule
    \textbf{Method} & \textbf{MedQA} & \textbf{MedXpert} & \textbf{MMLU-Pro} \\
    \midrule
    Raw margin       & 0.781 & 0.506 & 0.891 \\
    Inverted margin  & 0.219 & 0.494 & 0.109 \\
    Rank-corrected   & 0.781 & 0.506 & 0.891 \\
    DR-weighted      & 0.781 & 0.503 & 0.160 \\
    DACC (full)      & 0.781 & 0.507 & 0.892 \\
    \bottomrule
  \end{tabular}
\end{table}

Table~\ref{tab:correction} shows the AUROC of each correction method. On MedXpertQA, the best improvement is +0.001 (DACC: 0.507 vs.\ 0.506 raw), which is negligible. Inversion and rank correction fail because AUROC $\approx$ 0.5 means ranks are essentially random. Density-ratio reweighting identifies domain shift (mean ratio 3.999) but cannot determine which MedXpertQA questions are harder for the LLM; all questions appear similarly out of distribution. On MedQA and MMLU-Pro, methods either preserve the existing signal (rank-corrected, DACC) or destroy it (inverted margin: 0.219 on MedQA, 0.109 on MMLU-Pro). A hyperparameter sweep over $\alpha \in \{1.5, 2.0, 3.0\}$ and $\lambda \in \{0.3, 0.5, 0.7\}$ yields AUROC 0.49--0.50 across all configurations, confirming that no hyperparameter setting can extract a meaningful routing signal.

\begin{table}[t]
  \centering
  \caption{Adaptive retrieval results on MedXpertQA with corrected signals. All methods converge to ``always retrieve'' as the optimal strategy ($\tau{=}1.00$, 100\% retrieval), confirming the corrected signals provide no useful routing information.}
  \label{tab:adaptive_corrected}
  \small
  \begin{tabular}{lccc}
    \toprule
    \textbf{Method} & \textbf{Best $\tau$} & \textbf{Best ACC} & \textbf{Retrieval \%} \\
    \midrule
    Raw margin       & 1.00 & 26.0\% & 100\% \\
    Inverted         & 1.00 & 26.0\% & 100\% \\
    Rank-corrected   & 1.00 & 26.0\% & 100\% \\
    DR-weighted      & 1.00 & 26.0\% & 100\% \\
    DACC (full)      & 1.00 & 26.0\% & 100\% \\
    \midrule
    Never Retrieve   & ---  & 14.8\% & 0\% \\
    Always Retrieve  & ---  & 26.0\% & 100\% \\
    \bottomrule
  \end{tabular}
\end{table}

Table~\ref{tab:adaptive_corrected} confirms that all correction methods converge to $\tau{=}1.00$ (100\% retrieval), identical to the always-retrieve baseline. The corrected signal still provides no useful routing information: the problem is signal absence, not miscalibration.

\begin{figure*}[t]
  \centering
  \includegraphics[width=0.95\textwidth]{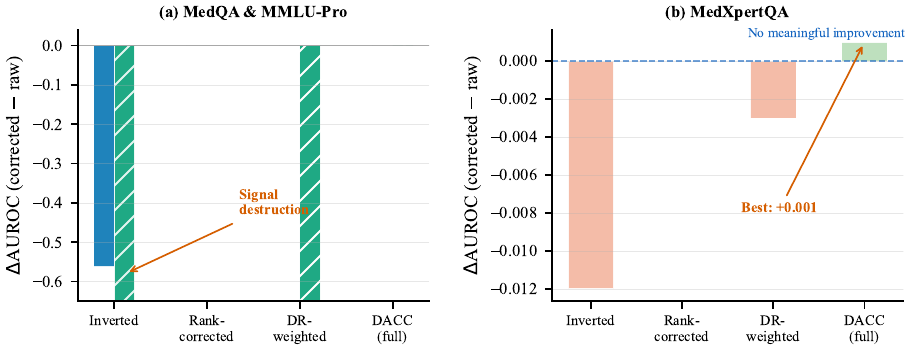}
  \caption{Effect of calibration correction, shown as $\Delta$AUROC (corrected minus raw margin). In panel (a), solid blue denotes MedQA and hatched green denotes MMLU-Pro. Inversion destroys the signal on easier datasets, while all methods produce negligible change on MedXpertQA (best: +0.001), confirming signal absence rather than miscalibration. Exact AUROC values in Table~\ref{tab:correction}.}
  \label{fig:correction}
\end{figure*}

\subsection{Cross-Model and Cross-Platform Verification}

To verify that the paradox is not an artifact of one model or platform, we replicate the experiment using the Alibaba DashScope API with verbalized confidence. We test five models spanning architectures, vendors, and parameter scales (229B--1.6T): Qwen3.7-Plus, DeepSeek-V4-Pro, Kimi-K2.6, GLM-5.1, and MiniMax-M2.5. Each model is evaluated on 200 samples per dataset with the same retrieval pipeline (MedQA training corpus, FAISS cosine similarity, top-5). This replication changes both the confidence extraction method and the serving infrastructure, providing a stringent robustness test.

\begin{figure*}[t]
  \centering
  \includegraphics[width=0.95\textwidth]{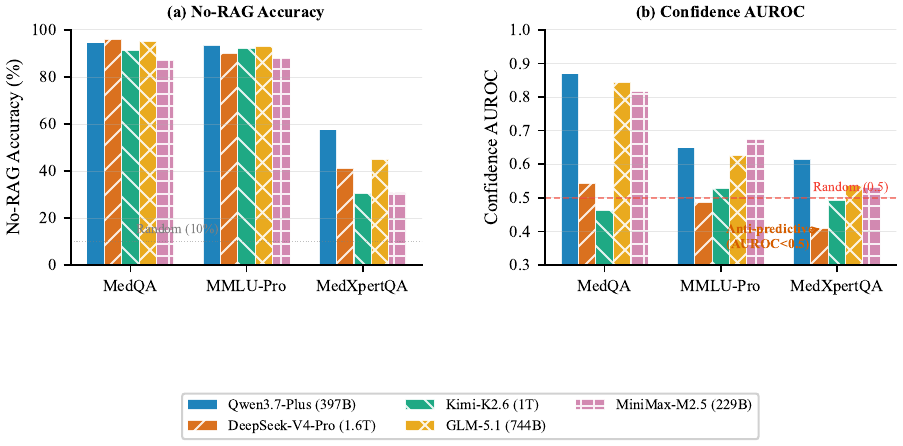}
  \caption{Cross-model verification across five models (DashScope API, verbalized confidence, 200 samples per dataset). Panels show no-RAG accuracy (a) and confidence AUROC (b). On MedXpertQA---the hardest dataset---DeepSeek-V4-Pro exhibits anti-predictive confidence (AUROC 0.408~$<$~0.5), while Kimi-K2.6 and MiniMax-M2.5 hover near chance (AUROC $\approx$~0.5). Qwen3.7-Plus retains the best calibration overall but sees AUROC drop from 0.869 (MedQA) to 0.614 (MedXpertQA). See Table~\ref{tab:cross_model} for exact values.}
  \label{fig:cross_model}
\end{figure*}

Table~\ref{tab:cross_model} summarizes the results.

\begin{table}[t]
  \centering
  \caption{Cross-model comparison using verbalized confidence (DashScope API, 200 samples per dataset). No-RAG accuracy and confidence AUROC are reported for each model. On standard datasets, larger models achieve high accuracy and moderate-to-strong calibration; on specialist-level MedXpertQA, confidence degrades to near-random for most models, confirming the paradox is not model-specific.}
  \label{tab:cross_model}
  \small
  \resizebox{\columnwidth}{!}{
  \begin{tabular}{llcccc}
    \toprule
    \textbf{Model} & \textbf{Params} & \textbf{Metric} & \textbf{MedQA} & \textbf{MMLU-Pro} & \textbf{MedXpert} \\
    \midrule
    \multirow{2}{*}{Qwen3.7-Plus} & \multirow{2}{*}{397B}
      & No-RAG ACC (\%)  & 94.5  & 93.5  & 57.5 \\
    & & Confidence AUROC & 0.869 & 0.649 & 0.614 \\
    \midrule
    \multirow{2}{*}{DeepSeek-V4-Pro} & \multirow{2}{*}{1.6T}
      & No-RAG ACC (\%)  & 96.0  & 90.0  & 41.0 \\
    & & Confidence AUROC & 0.542 & 0.485 & 0.408 \\
    \midrule
    \multirow{2}{*}{Kimi-K2.6} & \multirow{2}{*}{1T}
      & No-RAG ACC (\%)  & 91.0  & 92.0  & 30.5 \\
    & & Confidence AUROC & 0.461 & 0.529 & 0.493 \\
    \midrule
    \multirow{2}{*}{GLM-5.1} & \multirow{2}{*}{744B}
      & No-RAG ACC (\%)  & 95.0  & 93.0  & 45.0 \\
    & & Confidence AUROC & 0.842 & 0.627 & 0.538 \\
    \midrule
    \multirow{2}{*}{MiniMax-M2.5} & \multirow{2}{*}{229B}
      & No-RAG ACC (\%)  & 87.0  & 88.0  & 31.0 \\
    & & Confidence AUROC & 0.815 & 0.674 & 0.532 \\
    \bottomrule
  \end{tabular}
  }
\end{table}

Three findings stand out. First, the paradox replicates across all five models: on MedXpertQA, DeepSeek-V4-Pro's confidence is \emph{anti-predictive} (AUROC 0.408 $<$ 0.5), Kimi-K2.6 hovers at chance level (AUROC 0.493), and even Qwen3.7-Plus drops from 0.869 on MedQA to 0.614 on MedXpertQA. Second, stronger answer accuracy does not guarantee reliable routing: Qwen3.7-Plus, GLM-5.1, and DeepSeek-V4-Pro all exceed 90\% no-RAG accuracy on MedQA and MMLU-Pro, yet all exhibit confidence degradation on MedXpertQA. Third, calibration quality is not explained by parameter count alone: MiniMax-M2.5 (229B) achieves AUROC 0.815 on MedQA, while DeepSeek-V4-Pro (1.6T) reaches only 0.542.

\section{Analysis}

\subsection{Difficulty-Dependent Calibration Degradation}

A key observation is that calibration quality degrades systematically with dataset difficulty. On MedQA (72.2\% no-RAG accuracy), AUROC is 0.781; on MMLU-Pro Medical (80.8\% no-RAG accuracy), AUROC reaches 0.891; on MedXpertQA (14.8\% no-RAG accuracy), AUROC collapses to 0.506. This pattern holds across all five confidence signals (Tables~\ref{tab:multisignal} and~\ref{tab:signals}) and across all five models (Table~\ref{tab:cross_model}).

This degradation is not simply a consequence of low accuracy. Even on MedXpertQA, the model answers 14.8\% of questions correctly---far above the 10\% random baseline. The issue is that confidence scores do not \emph{discriminate} between correct and incorrect answers: the model is equally (un)confident on questions it gets right and wrong. This is qualitatively different from miscalibration (where confidence is systematically too high or too low but still discriminative); it represents a complete absence of discriminative signal.

The cross-model results sharpen this point. DeepSeek-V4-Pro (1.6T) achieves 41.0\% accuracy on MedXpertQA, nearly triple Qwen2.5-32B's 15.0\%, yet its confidence is anti-predictive (AUROC 0.408). Qwen3.7-Plus reaches 57.5\% accuracy and AUROC 0.614, the best calibration among the five models, but still below the discriminative power needed for dependable routing.

\subsection{Why Confidence Fails on Specialist-Level Questions}
Three factors help explain near-random calibration on specialist-level questions. First, MedXpertQA covers specialist knowledge where the LLM may lack the meta-cognition to recognize its own ignorance \cite{kadavath2022language}. Second, the 10-choice format lets the model concentrate probability on plausible but wrong options, amplifying the mismatch between confidence and correctness. Third, confidence reflects familiarity with general medical knowledge, which is poorly correlated with the expert reasoning required by MedXpertQA.

\paragraph{Why Calibration Correction Fails}
The failure of five correction methods reveals a deeper issue than miscalibration. AUROC $\approx$ 0.506 means the margin has essentially no discriminative power, so reweighting, inversion, or monotone transformation cannot recover a signal. Density-ratio estimation distinguishes MedQA from MedXpertQA questions (mean ratio 3.999), but domain shift alone does not identify which target questions the LLM will answer correctly. Rank-preserving correction \cite{hu2026rapcal} likewise fails because its core assumption---informative relative ordering under shift---is violated when ranks are near random.

\paragraph{Implications for Clinical AI Safety}
Our findings have direct consequences for retrieval-augmented clinical AI systems. Confidence alone is insufficient for routing in high-stakes biomedical domains: across five signals and five correction methods, none reliably predicts where retrieval helps on the hardest dataset. Validation on easy benchmarks can therefore overstate routing reliability. In clinical settings, always-retrieve (26.0\%) substantially outperforms confidence-based adaptive retrieval on MedXpertQA, and the cost of unnecessary retrieval is often lower than the cost of missing critical information \cite{topol2019high}. Safer systems need routing signals that do not depend solely on the LLM's self-assessment, such as external difficulty models, corpus coverage estimators, or task-specific domain classifiers.

\paragraph{Practical Recommendations}
Practitioners should benchmark adaptive routing on the hardest available dataset, not only on standard exams. Confidence AUROC below 0.55 should trigger skepticism about confidence-based routing. When target-domain calibration is unknown, a conservative policy is to retrieve by default or use hybrid routing: confidence-based decisions in validated domains and retrieval-on behavior in difficult or unverified domains.

\section{Conclusion}

We identify a paradox in confidence-based adaptive RAG for biomedical QA: retrieval is most valuable precisely where confidence fails as a routing signal. Across three datasets, confidence AUROC drops to 0.506 where retrieval offers $+11.2\%$. Five uncertainty signals and five calibration correction methods fail to recover a useful signal (best AUROC: 0.507), indicating signal absence rather than simple miscalibration. Cross-model replication across five DashScope models confirms the pattern. For high-stakes clinical applications, defaulting to always-retrieve is more reliable than confidence-based routing when target-domain calibration is unverified; developing non-confidence routing signals should therefore be a priority for safe clinical AI deployment.

\section*{Limitations}

Our main experiments use Qwen2.5-32B-Instruct, with five additional DashScope models for cross-model verification. Evaluations on more model families, retrieval corpora, and retrieval algorithms would further test generality. We evaluate biomedical multiple-choice QA; the same pattern may or may not extend to open-ended clinical reasoning or other expert domains. The cross-platform verification uses verbalized confidence rather than logprobs because of API constraints, introducing a methodological difference, although the consistent degradation pattern supports the core finding.

\bibliographystyle{IEEEtran}
\bibliography{references}

\end{document}